\documentclass[runningheads]{llncs}

\usepackage{graphicx}
\usepackage{amsmath,amssymb} 
\usepackage{color}
\usepackage[width=122mm,left=12mm,paperwidth=146mm,height=193mm,top=12mm,paperheight=217mm]{geometry}
\usepackage{epsfig}
\usepackage{epstopdf}
\usepackage{comment}
\usepackage{caption}
\usepackage{url}
\usepackage{array}
\usepackage[table]{xcolor}
\usepackage{ctable} 
\usepackage{subfig}
\usepackage{authblk}

\mainmatter
\def\ECCV16SubNumber{1616}  

\title{Reading Between the Pixels: Photographic Steganography for Camera Display Messaging}

\titlerunning{Reading Between the Pixels}

\authorrunning{Eric Wengrowski, Kristin Dana, Marco Gruteser, Narayan Mandayam}

\author{Eric Wengrowski\thanks{eric.wengrowski@rutgers.edu}
\and
Kristin Dana\thanks{kdana@rutgers.edu}
\and
Marco Gruteser\thanks{gruteser@winlab.rutgers.edu}
\and
Narayan Mandayam\thanks{narayan@winlab.rutgers.edu}}

\institute{Rutgers University\\ Department of Electrical and Computer Engineering\\ WINLAB (Wireless Information Network Laboratory)\\ Piscataway, NJ}

\begin{document}

\maketitle
\thispagestyle{empty}

\begin{abstract}
We exploit human color metamers to send light-modulated messages less visible to the human eye, but recoverable by cameras. These messages are a key component to camera-display messaging, such as handheld smartphones capturing information from electronic signage. 
Each color pixel in the display image is modified by a particular color gradient vector. The challenge is to find the color gradient that maximizes camera response, while minimizing human response.
The mismatch in human spectral and camera sensitivity curves creates an opportunity for hidden messaging. Our approach does not require knowledge of these sensitivity curves, instead we employ a data-driven method. 
We learn an ellipsoidal partitioning of the six-dimensional space of colors and color gradients. This partitioning creates metamer sets defined by the base color at the display pixel and the color gradient direction for message encoding. 
We sample from the resulting metamer sets to find color steps for each base color to embed a binary message into an arbitrary image with reduced visible artifacts. 
Unlike previous methods that rely on visually obtrusive intensity modulation, we embed with color so that the message is more hidden. 
Ordinary displays and cameras are used without the need for expensive LEDs or high speed devices.
The primary contribution of this work is a framework to map the pixels in an arbitrary image to a metamer pair for steganographic photo messaging.
\end{abstract}

\section{Introduction}

\begin{figure}[!t]
\centering
\includegraphics[width=0.75\textwidth]{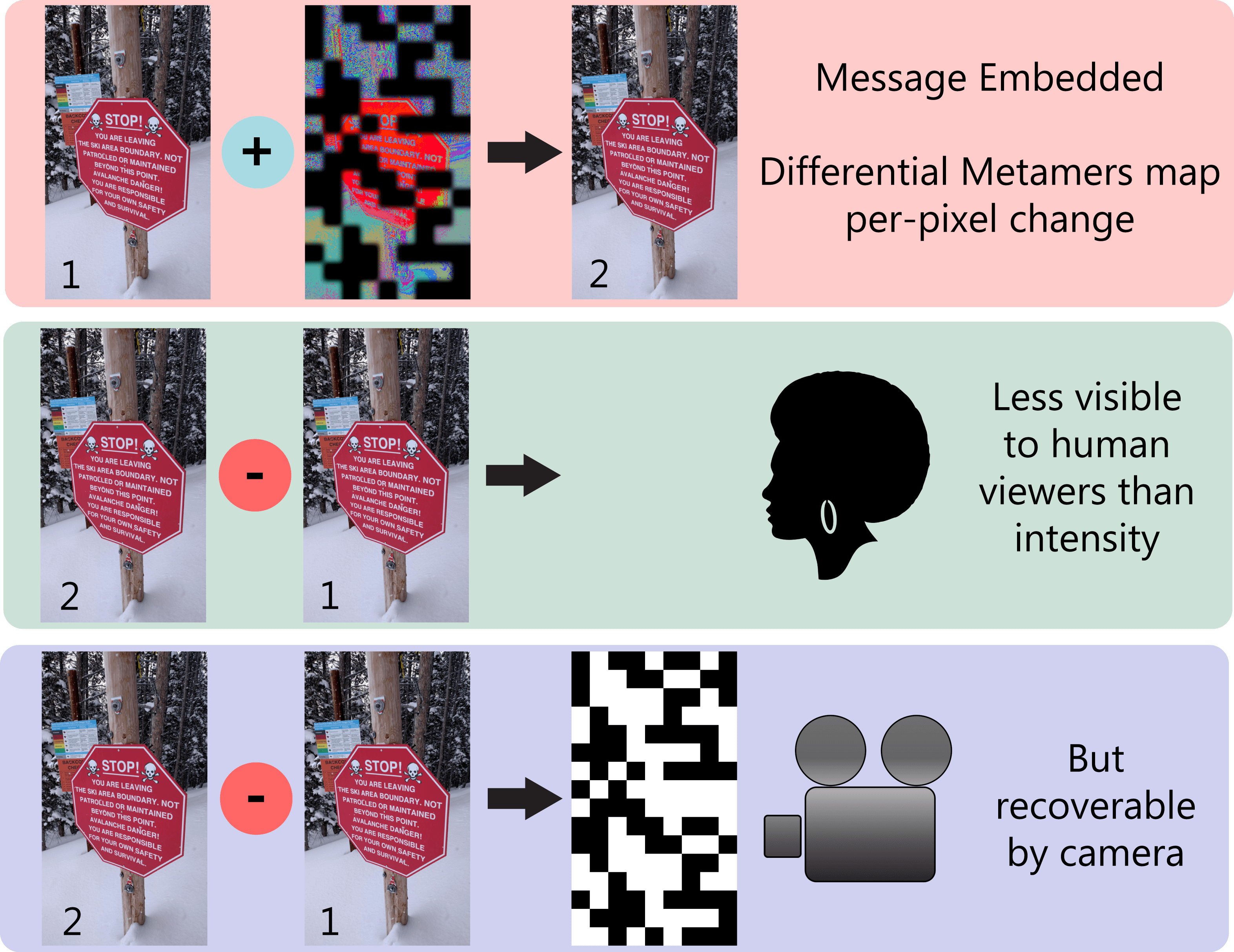}
\caption{Differential Metamers are color pairs that are optimized to be hidden from human vision, but sensitive to a camera. By modulating a small, per-pixel change in an image, differential metamers can be used to embed hidden messages. As shown above, the embedded message is blended to reduce the spacial visibility without disturbing camera recovery.}
\label{fig:opening_fig}
\end{figure}

Electronic displays, such as LCD monitors, are typically used for the single task of human visual observation.
Research in the relatively new field of camera-display communication has introduced a dual channel: a machine-readable communications channel operating in parallel with the human-observable display.
Time-varying messages can be embedded in the on-screen images, however the task has significant challenges.
The modulated signal is an illumination field propagating in free-space, so prior methods of watermarking for digital images are not directly applicable. The illumination field emitted by the display and captured by the camera depends on the parameters of the radiometric transfer function and sensitivity curves of both the display and camera.  
This camera-display transfer function makes message recovery challenging, but it also presents an opportunity for message embedding that is  tuned to typical transfer functions.

A common method for camera-display messaging relies on intensity modulation either for directly embedding bits or for embedding transformation coefficients~\cite{Wengrowski2016optimal}.
Human vision is generally very sensitive to intensity step edges, even when the step size is small. 
For simple messaging, the display image can be modified by adding  a message image where ``1'' bit values are encoded in a block by a small  $\delta$ value  and ``0'' bit values are encoded by a zero block value. The message frame is added in alternative temporal frames so that 
sequential frame subtraction can be used to decode the message.
This method assumes that  the display image is constant over time intervals. Accurate message recovery is challenging because small $\delta$ values are needed to hide the message, but  large $\delta$
values are needed for a low-noise signal that can be accurately decoded by the camera. 
%
Another approach to making the message imperceptible is to use high speed light modulation so that the flicker fusion effect of human vision can temporally blur the intensity variation.  High speed displays are commercially available, but their higher cost can be prohibitive in electronic signage and mobile display applications. 

Our approach uses {\it color message
 embedding} that exploits the differences in human color sensitivity versus camera color sensitivity. This method has a key advantage of not needing  specialized hardware. 
For color embedding the display image $i(x,y)$ has 3 components ($i_r, i_g,i_b$) and a color message image $m(x,y)$ is added to the display image. The goal is to find the best color direction for $m(x,y) = \delta \hat{c}$ where $\hat{c}$ is a unit direction in color space and $\delta$ is the step-size. 
We seek a {\it differential metamer} ($i$,$i+\delta\hat{c}$) such that $i+\delta \hat{c}$ is perceived to be the same color as $i$ by a human observer but is captured by the camera as a distinguishable color. 
The differential metamer is comprised of a three component base color and three component direction vector in color space.
The key innovation is a method for finding differential metamers. Our approach uses a  six-dimensional quadratic binary classifier, solved in a convex optimization problem. Using training data with positive and negative examples, the algorithm determines a set of ellipsoids in six dimensional space such that the 
interior of these ellipsoids produce points $p$ where the first three components corresponds to  a particular base color pixel $i$ and the next three components  provide the  color step direction $\hat{c}$ to use for  messaging for a particular base color pixel.
The interior of these six dimensional ellipsoids define approximate metamer sets that are sufficiently accurate to provide message hiding 
while supporting message recovery. 

\subsection*{Differential Metamers}
We  introduce the term {\it differential metamers} to define pairs of color values programmed for display that result in minimal visible change for the human observer when viewed sequentially, but are distinguishable colors when captured by a camera. This process is illustrated in Figure~\ref{fig:opening_fig}.
Traditionally, the term {\it metamers} are colors that have different spectral power distributions, but appear identical to human observers.
Differential metamers are defined with respect to the color display as a pair of 3-component vectors representing the display's programmable colors that appear identical to human observers but significantly different to the camera.  
Many differential metamers exist even among 8-bit color values, but finding the color values that yield both low human sensitivity and high camera sensitivity is difficult because over $10^{14}$ colors would need to be tested for both camera sensitivity and human sensitivity.
Specific camera sensitivity curves combined with human vision parameters would not be enough to model the differential metamer space. Display parameters indicating the spectrum  of light emission for each programmed color vector and the dependence on camera observation parameters would also be needed to determine an analytical model.
Given the variations involved, we choose a data driven approach instead. We show that this approach is straightforward and effective.  By sampling $~10^{3}$ points, we can train a union of ellipsoidal binary classifiers to predict successful differential metamers where the base color values $c$ can lie in the full color space. We perform the metamer set estimation in both  RGB and  CIE $Lab$ color spaces.
 

\section{Background and Related Work}

\begin{figure}[t]
\centering
\includegraphics[width=0.5\textwidth]{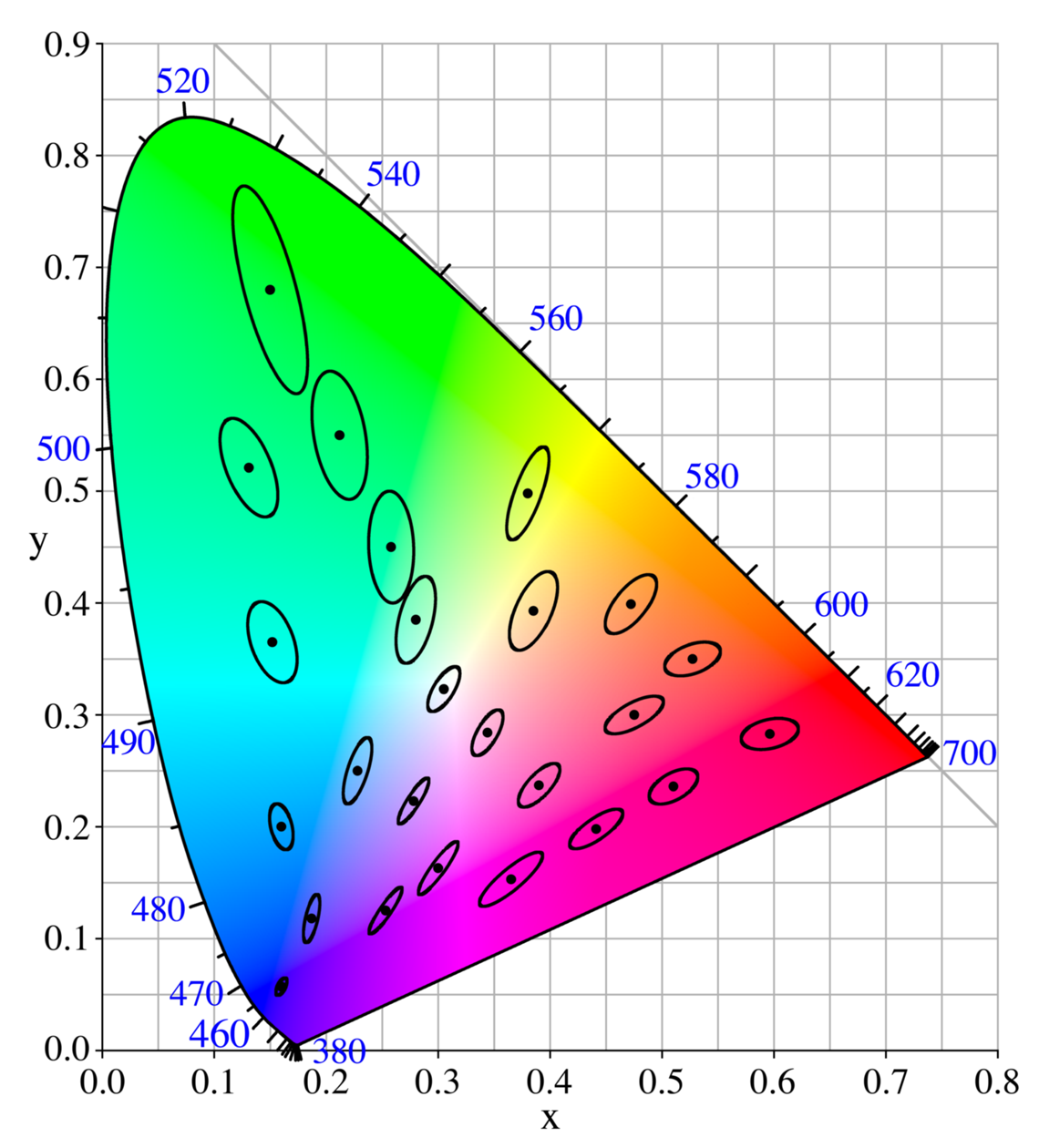}
\caption{MacAdam ellipses for the CIE xy 1931 colorspace \cite{macadam1943specification,1_wikipedia_2015}.  The area within these scaled-up ellipses represent metamers, or colors which cannot be distinguished.}
\label{fig:macadams}
\end{figure}

\paragraph{Metamers and Separating Ellipsoids}
Our approach to finding separating ellipsoids in color space is motivated by two main factors. First, the problem of fitting a separating ellipsoid to labeled data is a convex optimization problem \cite{boyd2004convex} and therefore is not affected by local minima.  Second, human vision research has showed the utility of ellipsoidal surface fitting for representing  color difference thresholds. 
As early as the 1940's, human vision studies identified and quantified ellipsoidal representations 
for the problem of understanding human sensitivity to small color differences \cite{brown1949visual,macadam1943specification} as illustrated in Figure~\ref{fig:macadams}.
This ellipsoidal representation was confirmed in numerous  studies in 
early vision literature \cite{wyszecki1971new,brown1957color,davidson1951calculation}. 
Parametric surfaces were used to find discrimination contours and the fitting typically used detection thresholds \cite{Poirson90,wyszecki2009color} in order to get just-noticable-difference JND contours \cite{Noorlander83}. Our framework greatly simplifies this process because no threshold values are measured. Instead, a learned separating ellipsoid finds a discrimination boundary between color pairs that are differential metamers and those that are not. 
Metamer sets \cite{finlayson2005metamer} are convex hulls, which ellipsoids are well-suited to fit.
By extension, we have adopted discriminating ellipsoids to characterize the space of differential metamers. 
In a related prior work for using    color to embed information \cite{rudaz2004protecting}  color gradients are used to watermark spatially varying microstructures into images. The objective in this work is to use color  to embed watermarks that were difficult to see from a distance, but visible up close.  This is different from our goal of finding pairs of colors where no distinction can be made when viewed sequentially by humans, but the difference can be robustly detected by a camera. 

\paragraph{Camera-Display Communication}
\indent Electronic displays such as televisions, computer monitors, and projectors are traditionally used to display images, videos, and text - all human readable scenes. These devices can also display machine-readable images such as QR-codes \cite{Wengrowski2016optimal,hu2014strata,hao2012cobra,yuan2012dynamic,yuan2014phase,li2014hilight,perli2010pixnet,jo2014disco}.
Within the past 5 years, extensive work has been done to expand the capabilities of camera display messaging by increasing throughput. 
\\
\indent PixNet introduced OFDM transmission algorithms to address the unique characteristics of the camera-display link, including perspective distortion, blur, and sensitivity to ambient light \cite{perli2010pixnet}. While PixNet offer impressive data throughput, it can only display machine-readable code and supports no hybrid approach. Strata introduced distance-scalable coding schemes \cite{hu2014strata}, preferable in a mobile application, but also cannot display both human-readable and machine readable images at the same time. Both of the aforementioned techniques encode bit values with intensity. COBRA introduced a 2D color code \cite{hao2012cobra}, but also could only display machine readable code.
\\
\indent Both Visual MIMO and HiLight employ intensity modulation of human-readable images to embed a second machine-readable channel \cite{DBLP:journals/corr/YuanWDAGM15,li2014hilight}. However, it is well known that human vision is extremely sensitive to temporal and spacial to changes in intensity. It has been shown that intensity changes, even with small magnitude are likely to cause flicker and discomfort to a human observer. The amount of human visual obtrusion had not been measured for either method.
\\
\indent Kaleido and VRCodes uses metamers to embed data in alternating pixel values \cite{zhang2015kaleido,woo2012vrcodes}. These values, however, are not ``true'' metamers in the sense that two static colors have different physical properties such as wavelength, but appear identical to human viewers. Instead, Kaleido and VRCodes leverages flicker fusion to create temporally blended colors hidden from human observers with high speed changes. This method is limited by the requirement for specialized high-speed displays. VRCodes also leverages the rolling shutter camera typically found on mobile phones to sample at frequencies above 60Hz. Unfortunately, this limits VRCode throughput to only 1 bit per frame.
\\
\indent
On the other hand, Kaleido attempts to solve a different problem: embedding noise with flicker fusion metamers to disrupt piracy via camera recording of videos, while preserving the human-visible channel. While similar in intuition to the work presented in this paper, the goals are fundamentally different.
We embed camera-sensitive information in this invisible channel, while Kaleido only embeds camera-sensitive noise. And as stated before, Kaleido requires specialized high-speed displays, while our method requires no specialized hardware.
\\
\indent LED arrays have used modulated light to communicate \cite{Ashok11,ashok2014not}. Recently, LED-based communication techniques have used color-shift keying for communication \cite{luoexperimental}. Methods exist to make this color-shift keying imperceptible to human observers \cite{hucolorbars}, but these applications do not require the imperceptible reproduction of entire images.
\\
\indent In this work, we take a data driven approach to generating differential metamers that have a small human sensitivity gradient, but high camera sensitivity gradient. We show that differential metamers are effective for embedding hidden messages into images on electronic displays. And we have derived an empirical model for the effectiveness of each RGB base color and corresponding step values for flicker-free embedding.


\section{Learning Differential Metamers}

Our approach to finding differential metamers for color embedding starts by creating a training set of base colors and color steps that meet the criterion for embedding (low human sensitivity and high camera sensitivity). The set of color pairs that are either visible when viewed sequentially or not detectable by the camera are negative training examples. The positive and negative examples reside in six dimensional space $\mathbb{R}^6$. We choose the number of separating ellipsoids $k$ empirically and cluster the positive examples into $k$ clusters in  $\mathbb{R}^6$. For each cluster, we find the optimal ellipsoid that separates positive and negative data.
Sampling within the union of all ellipsoids, reveals a dense set of  new differential metamers.

\subsection{Training $k$ Optimal Separating Ellipsoids}

\begin{figure}[h]
	\centering
	\includegraphics[width=0.75\textwidth]{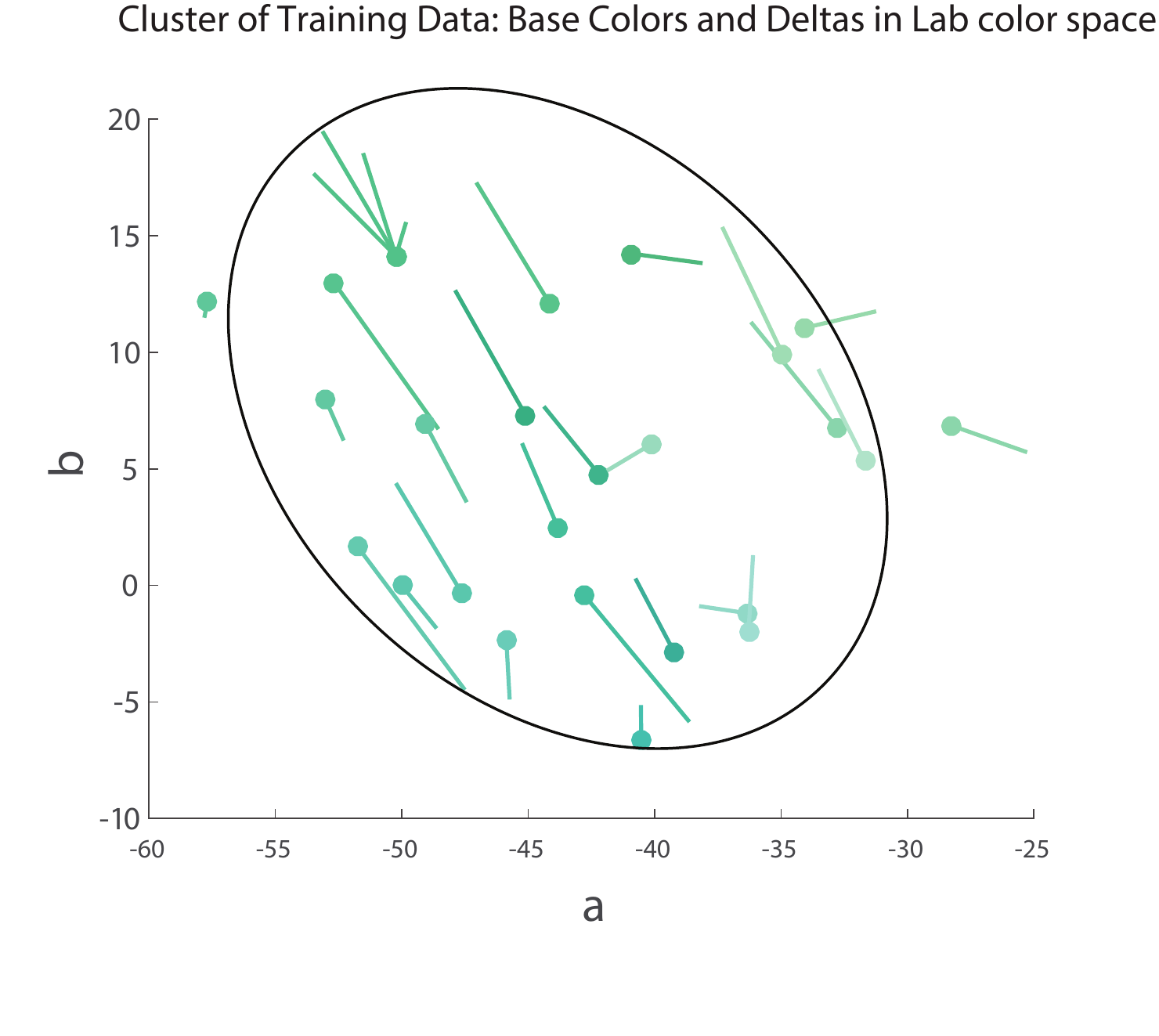}
    \vspace{-1cm}
	\caption{ In this figure, a separating ellipsoid and cluster of positively labeled training data have been projected down to $a$ $b$ space (from $Lab$ color space). Visualizing $k$ 6-dimensional ellipsoids is difficult. Instead, we omit plotting the $L$ dimension and show the base colors and color shifts at the same time for a single training cluster. The solid circle represents the base color, and its respective line segments represents the color shift. The color of each circle and line segment is the actual base color. Notice how there is a general axis of color shift direction for the data in this cluster. Since these are positive training points, this indicates that human viewers are relatively insensitive to these relative color shifts. This also indicates that our camera is sensitive to these relative color shifts. }
\label{fig:ellipsoid}
\end{figure}

For each cluster $k_i$, the optimal separating ellipsoid is found. Each ellipsoid separates the positive training points in cluster $k_i$ from all negative training points.
\\
\indent
We have two sets of points in $\mathbb{R}^6$, $\{x_1,...x_N\}$ and $\{y_1,...y_M\}$. 
The points $x_i$ represent the base colors and modulation steps that satisfy the requirements for embedding: BER $= 0\%$, and no visible flicker. 
While the points $y_i$ do not satisfy both of these conditions. We wish to find a function $f:\mathbb{R}^n \to \mathbb{R}$ that is positive on the first set, and negative on the second, \textit{i.e.},

\begin{equation}
f(x_i)>0, \  i = 1,...,N, \qquad f(y_i)>0, \  i = 1,...,M. 
\end{equation}
When these inequalities hold, we say that $f$ separates the two sets of points.


\paragraph{Quadratic Discrimination}

Since our data points cannot be separated by a $N$-dimensional hyperplane, we seek classification via nonlinear discrimination. 
As long as the parameters that define $f$ are linear (or affine), the above inequality can still be solved with convex optimization.
\\
\indent
In this case, we choose $f$ to be quadratic and in homogeneous form:
\begin{equation}
f(z)=z^TPz + q^Tz + r,
\end{equation}
where $P \in \textbf{S}^n$ ($P$ is a symmetric $n\times n$ matrix), $\ q \in \mathbb{R}^n,$ and $ r \in \mathbb{R}$, with dimensionality n = 6. Those parameters $P, \ q, \ r$ are bound by the following constraints:

\begin{equation}
\begin{split}
\begin{aligned}
x_i^TPx_i + q^Tx_i + r &> 0, \qquad &&i = 1,...,N
\\
y_i^TPy_i + q^Ty_i + r &< 0, \qquad &&i = 1,...,M
\end{aligned}
\end{split}
\label{eq:ineq0}
\end{equation}

Notice the sign and direction of the inequalities in Eq.~\ref{eq:ineq0} seem counter intuitive, but it is correct. Next, we replace $0$ with $\epsilon$, creating a separating band that is $2\epsilon$ wide:

\begin{equation}
\begin{split}
\begin{aligned}
x_i^TPx_i + q^Tx_i + r &\geq  \epsilon, \qquad &&i = 1,...,N
\\
y_i^TPy_i + q^Ty_i + r &\leq -\epsilon, \qquad &&i = 1,...,M
\end{aligned}
\end{split}
\label{eq:ineq_ep}
\end{equation}

Dividing out by $\epsilon$ and subsuming the scalar $\frac{1}{\epsilon}$ into $P,q,r$, you arrive at Eq.~\ref{eq:ineq1}. Following the Boyd text \cite{boyd2004convex}, we solve for the parameters $P, \ q, \ r$ by solving the non-strict feasibility problem:

\begin{equation}
\begin{split}
\begin{aligned}
x_i^TPx_i + q^Tx_i + r &\geq  1, \qquad &&i = 1,...,N
\\
y_i^TPy_i + q^Ty_i + r &\leq -1, \qquad &&i = 1,...,M
\end{aligned}
\end{split}
\label{eq:ineq1}
\end{equation}

The resulting separating surface $\{z \, | \, z^TPz + q^Tz + r=0\}$ is quadratic. 

\subsubsection*{Separating Ellipsoids}

We can change the shape of our quadratic separating surface by imposing additional constraints on the parameters $P, \ q,$ and $ r$. 
We form an ellipsoid that contains all points $x_i,...,x_N$ but none of the points $y_i,...,y_M$ by requiring that $P\prec 0$, that is $P$ is negative semi-definite. 
We can use homogeneity in $P, \ q, \ r$ to express the constraint $P\prec 0$ as $P \preceq -I$. 
We can then cast our quadratic discrimination problem as the following semi-define programming (SDP) feasibility problem:

\begin{equation}
\begin{split}
\begin{aligned}
&\text{find} && P, \ q, \ r
\\
&\text{subject to} && x_i^TPx_i + q^Tx_i + r \geq 1, \qquad &&i = 1,...,N
\\
& && y_i^TPy_i + q^Ty_i + r \leq -1, \qquad &&i = 1,...,M
\\
& && P \preceq -I
\end{aligned}
\end{split}
\end{equation}

While technically correct, this optimization problem will fail if any of the training points fall outside their classification boundaries. 
Instead, we borrow a technique from support vector classifiers, and relax our constraints by introducing non-negative variables $u_1,...,u_N$ and $v_1,...,v_M$. 
With the relaxation variables $u_i$ and $v_i$ introduced, our inequalities become:
\begin{equation}
\begin{split}
\begin{aligned}
x_i^TPx_i + q^Tx_i + r &\geq 1-u_i, \qquad &&i = 1,...,N
\\
y_i^TPy_i + q^Ty_i + r &\leq v_i-1, \qquad &&i = 1,...,M
\\
\end{aligned}
\end{split}
\end{equation}

The relaxation variables $u_i$ and $v_i$ represent the distances of each point outside it's proper boundary. In the original problem, $u = v = 0$. We can think of  $u_i$ as a measure of how much each constraint $x_i^TPx_i + q^Tx_i + r \geq 1$ is being violated and that's what we want to minimize. 
A good heuristic is minimizing the sum of variables $u_i$ and $v_i$.
We solve for our supporting ellipsoid defined by $P, \ q, \ r$ with the following optimization problem:
\begin{equation}
\begin{split}
\begin{aligned}
&\text{minimize} && \textbf{1}^Tu + \textbf{1}^Tv
\\
&\text{subject to} && x_i^TPx_i + q^Tx_i + r \geq 1-u_i, \qquad &&i = 1,...,N
\\
& && y_i^TPy_i + q^Ty_i + r \leq v_i-1, \qquad &&i = 1,...,M
\\
& && P \preceq -I
\\
& && u \preceq 0, \qquad v \preceq 0
\end{aligned}
\end{split}
\end{equation}
To solve this problem we used CVX, a package for specifying and solving convex programs \cite{cvx,gb08}. After each ellipsoid is solved, we test that the ellipsoid is populated before accepting it.

\subsection{Sampling Within Union of Ellipsoids}
Once $k$ optimal separating ellipsoids are trained, the points inside the ellipsoid reflect desirable values for message embedding. So, to expand our gamut of differential metamers, we densely sample inside the ellipsoid region for new points.

\section{Experiments}

\subsubsection*{Collecting Training Data}

Base colors are generated by randomly sampling Lab space. For each base color, 20 baricentrically sampled unit vectors are generated. These unit vectors represent color shifts in Lab space along a particular direction.
Each data point is cast as a 6-dimensional vector $g$, such that: 
\begin{equation}
g = [c_L,\  c_a, \ c_b, \ c_L+d_L, \ c_a+d_a, \ c_b+d_b]^T.
\end{equation}
$c_L, c_a, c_b$ are the Lab values of the base color, and $d_L, d_a, d_b$ is the Lab color shift. The magnitude of the message step-size is defined as the Euclidean distance:
\begin{equation}
|\delta|_2 = \sqrt[]{d_L^2 + d_a^2 + d_b^2}
\end{equation}

\subsubsection*{Labeling Training Data}

A video sequence is generated as follows: Odd frames consisted of only a monochromatic image of the base color. Even frames comprised of the base color and a 2D barcode grid, 16 blocks wide and 9 blocks tall. Each block in this grid corresponded to 1 bit, with 144 bits per frame in total. No change to the base color keys a ``0'' bit. A camera-detected color shift keys a ``1'' bit. The Lab ``distance'' of the color shift corresponds to the radius of each ray, scaled by a constant. Here, the Lab shift radius was 5 (on the 8-bit [0 255] scale). All ``1'' blocks in a single frame are color shifted in the same direction, with the same radius.
\\
\indent
Positive points are defined as ones whose color embedding is invisible to humans, but recoverable by camera with BER (bit error rate) $= 0\%$. All other point pairs are negative training data. After labeling, 922 positive and 1558 negative data points were used for training. Empirically, we chose the number of clusters $k = 50$.

\subsubsection*{Evaluation of Clustering Methods}

\begin{figure}[!h]
\centering
  \begin{tabular}{ |l|r|r|r|r|r| }
   \hline
   \multicolumn{1}{|l|}{\begin{tabular}[x]{@{}l@{}} Clustering \\ Algorithm \end{tabular}}
   & \multicolumn{1}{l|}{\begin{tabular}[x]{@{}l@{}} Low Exposure \\ Mean Error \end{tabular}}
   & \multicolumn{1}{l|}{\begin{tabular}[x]{@{}l@{}} Low Exposure \\ STD \end{tabular}}
   & \multicolumn{1}{l|}{\begin{tabular}[x]{@{}l@{}} High Exposure \\ Mean Error \end{tabular}}
   & \multicolumn{1}{l|}{\begin{tabular}[x]{@{}l@{}} High Exposure \\ STD \end{tabular}}
   & \multicolumn{1}{l|}{\begin{tabular}[x]{@{}l@{}} Runtime \\ \textit{(sec)}\end{tabular}} \\
   \specialrule{0.5pt}{0pt}{0pt}
  $k$-Means & 30.41\% & 10.65\% & 24.16\% & 11.08\% & 0.0435 \\
  \specialrule{0.5pt}{0pt}{0pt}
  $k$-Medoids & 28.97\% & 10.89\% & 22.97\% & 9.92\% & 0.5813 \\
  \specialrule{0.5pt}{0pt}{0pt}
  \begin{tabular}[x]{@{}l@{}} Gaussian  \\ Mixture \\ Models \end{tabular} & 27.33\% & 10.83\% & 22.72\% & 11.05\% & 0.0978 \\
  \specialrule{0.5pt}{0pt}{0pt}
  \begin{tabular}[x]{@{}l@{}} Hierarchical \\ clustering \end{tabular} & 29.37\% & 11.42\% & 22.97\% & 11.64\% & 0.1299 \\
  \specialrule{0.5pt}{0pt}{0pt}
  \begin{tabular}[x]{@{}l@{}} Spectral \\ clustering \end{tabular} & 34.52\% & 11.58\% & 24.70\% & 9.91\% & 0.1387 \\
  \specialrule{0.5pt}{0pt}{0pt}
  \end{tabular}
\vspace{5pt}
\caption{ Camera recovery error for various clustering methods (\textit{lower is better}). Gaussian Mixture Models (GMMs) produce results with the lowest average errors under both exposure conditions. In this case, GMMs provide a great balance of runtime cost and performance. 
}
\label{tab:results_chart}
\end{figure}

To determine which clustering algorithm to use, differential metamers were trained and learned using each of the clustering methods in Table~\ref{tab:results_chart}. Then, these differential metamers were used for message embedding, and the messages were camera-recovered. The evaluation procedure is detailed in Section 4.
This evaluation is performed twice for each clustering algorithm under two different illumination conditions. Once where the camera has fixed high-exposure settings, and once again with fixed low-exposure settings. 
Under both exposure conditions, Gaussian Mixture Models (GMMs) produce embedding with lower recovery errors on average. Although the margin of superiority was small, GMMs were ultimately chosen. Regardless of method used or illumination condition, the standard deviation hovered around $10\%$ for all methods. This suggests that the recovery error results are largely dependent on the base image used. This result has been verified empirically as well; certain images produce better embedding results.
The run time calculations took place on an Intel 6700K processor with 32 GB of memory running Matlab 2015b.

\subsection*{Embedding Photo Steganographic Messages}

The message structure we employ is a 2D barcode grid, 16 blocks wide and 9 blocks tall, containing 144 bits in total.
Once a set of differential metamers is generated, these color pairs can be used to embed messages into ordinary images. For the $l^{th}$ pixel $p$ in an image $i$, we have $p^{(i,l)}$. For $p^{(i,l)}$, we compute the nearest base color in the set of differential metamers, $c^{(j)}$. If $p^{(i,l)}$ belongs to a block keyed with a ``1'' bit, then the corresponding $\delta$, $d^{(j)}$, is added to $p^{(i,l)}$. When the $l^{th}$ pixel is displayed in a video sequence, frame(1) will contain $p^{(i,l)}$ and frame(2) will contain $p^{(i,l)} + d^{(j)}$.

\subsection*{Recovering Photo Steganographic Messages}

\indent A camera stationed approximately 0.5 meters from an electronic display captured each pair of images. The camera had a fixed shutter speed, ISO sensitivity, aperture, and white balance. The images used were a monochromatic base image $i$, of color
\[
g_{base}^{(i)} = ( c_{L}^{(i)}, c_{a}^{(i)}, c_{b}^{(i)} ),
\]
and the embedded message was of color
\[
g_{message}^{(i)} = ( d_{L}^{(i)}, d_{a}^{(i)}, d_{b}^{(i)} ).
\]
The two frames are camera-captured, and then subtracted from each other. For each bit-block, an average difference greater than some threshold corresponds to a ``1'', and below that threshold corresponds to a ``0''. The recovered message was then compared to the known message to calculate BER (bit error rate). BER is the percentage of errors in each 144 bit message.

\subsubsection*{Testing Recovered Bit Error Rate (BER)}

\begin{table}[t]
\hspace{-0.25cm}
  \begin{tabular}{ ccccccc }
    \begin{minipage}{0.14\textwidth}
      \includegraphics[width=\textwidth]{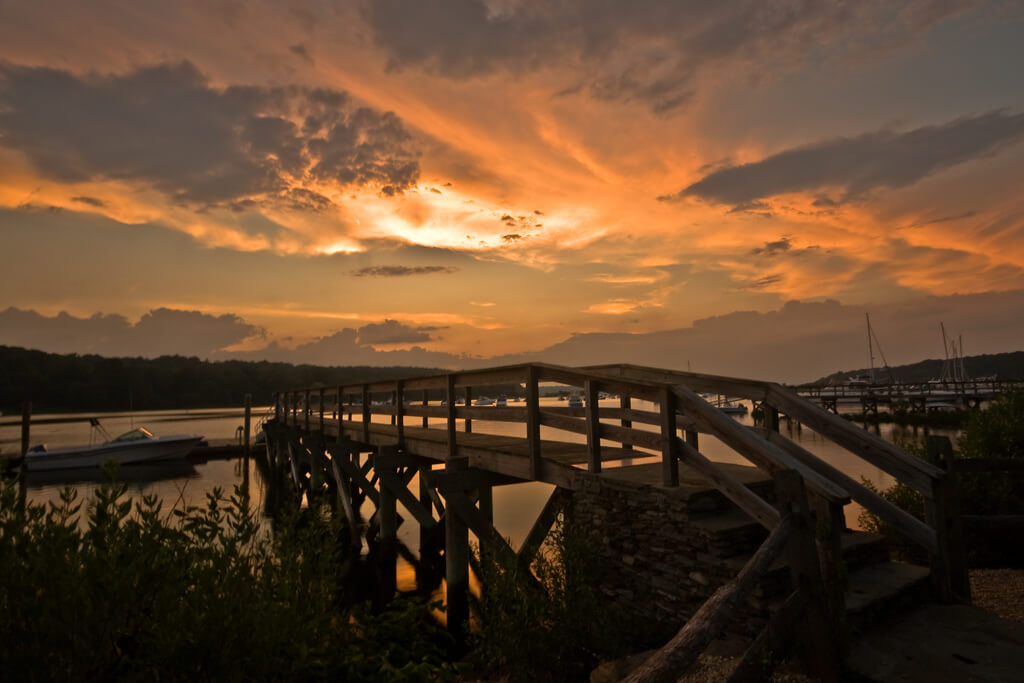}
    \end{minipage}
    &
    \begin{minipage}{0.14\textwidth}
      \includegraphics[width=\textwidth]{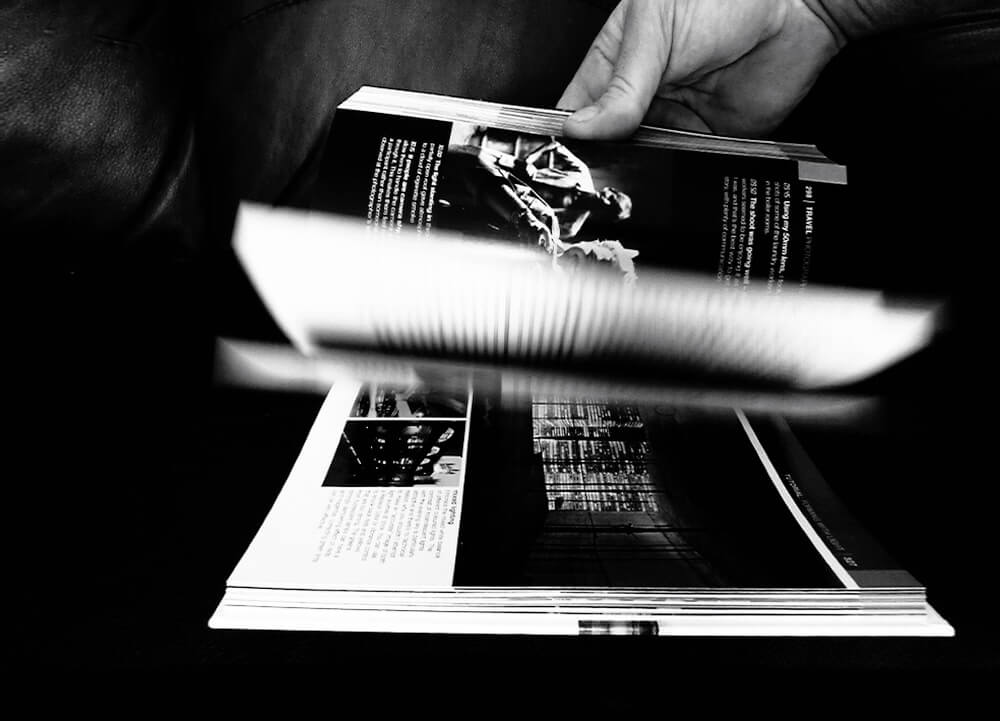}
    \end{minipage}
    & 
    \begin{minipage}{0.14\textwidth}
      \includegraphics[width=\textwidth]{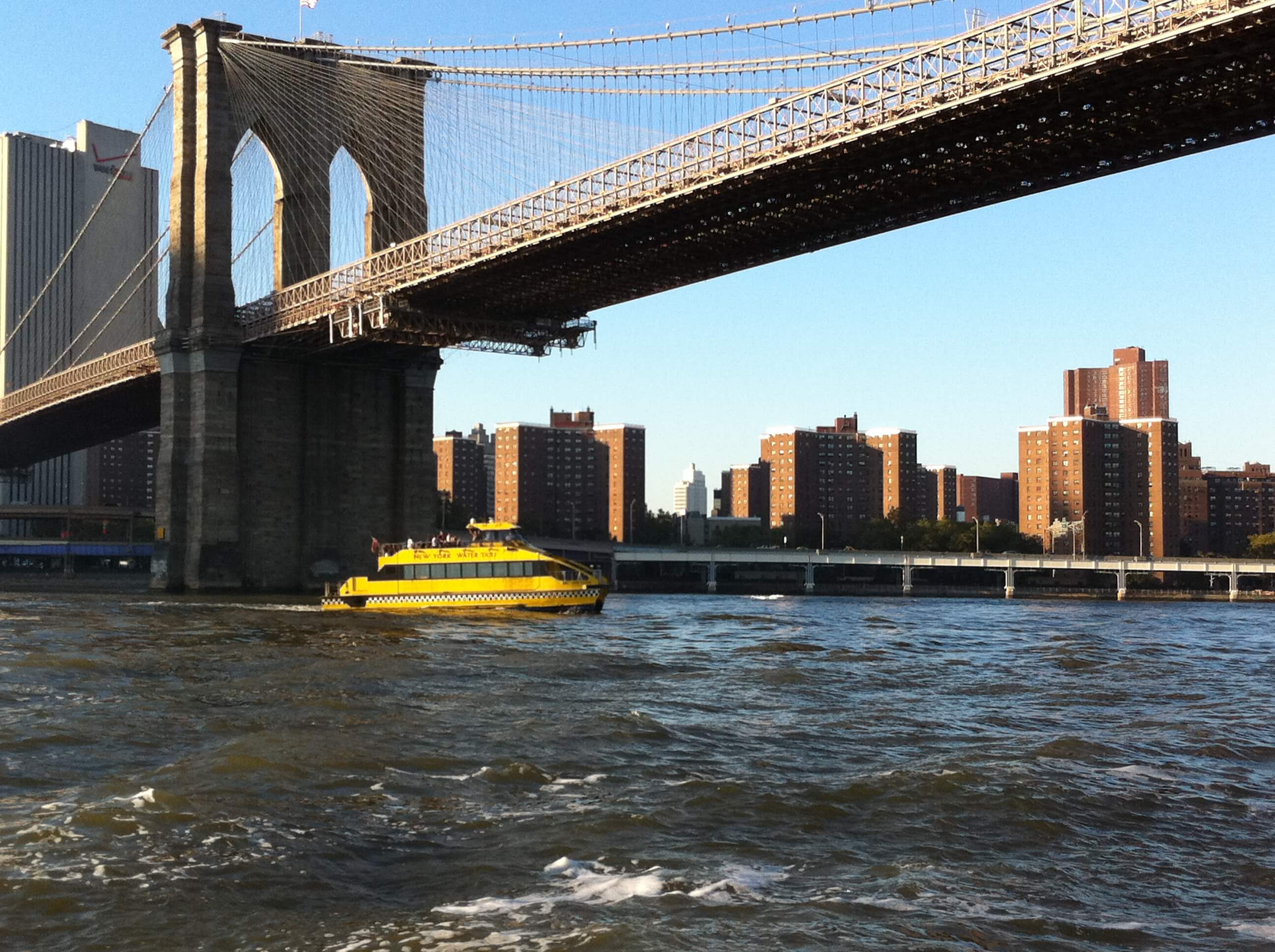}
    \end{minipage}
        & 
    \begin{minipage}{0.14\textwidth}
      \includegraphics[width=\textwidth]{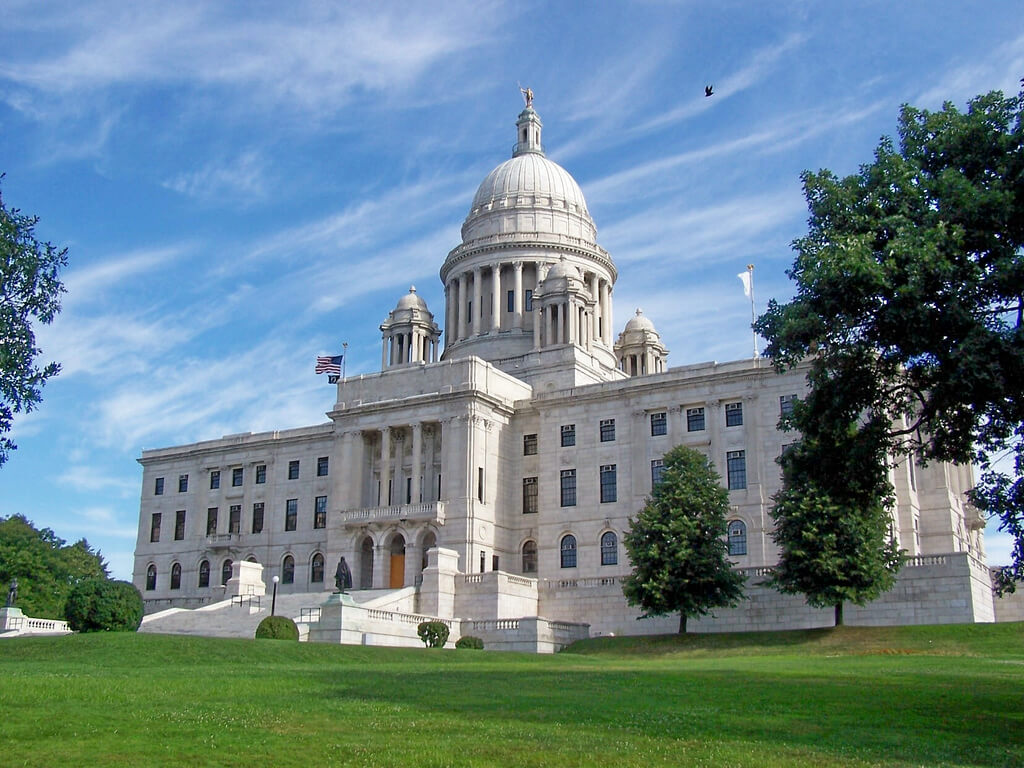}
    \end{minipage}
        & 
    \begin{minipage}{0.14\textwidth}
      \includegraphics[width=\textwidth]{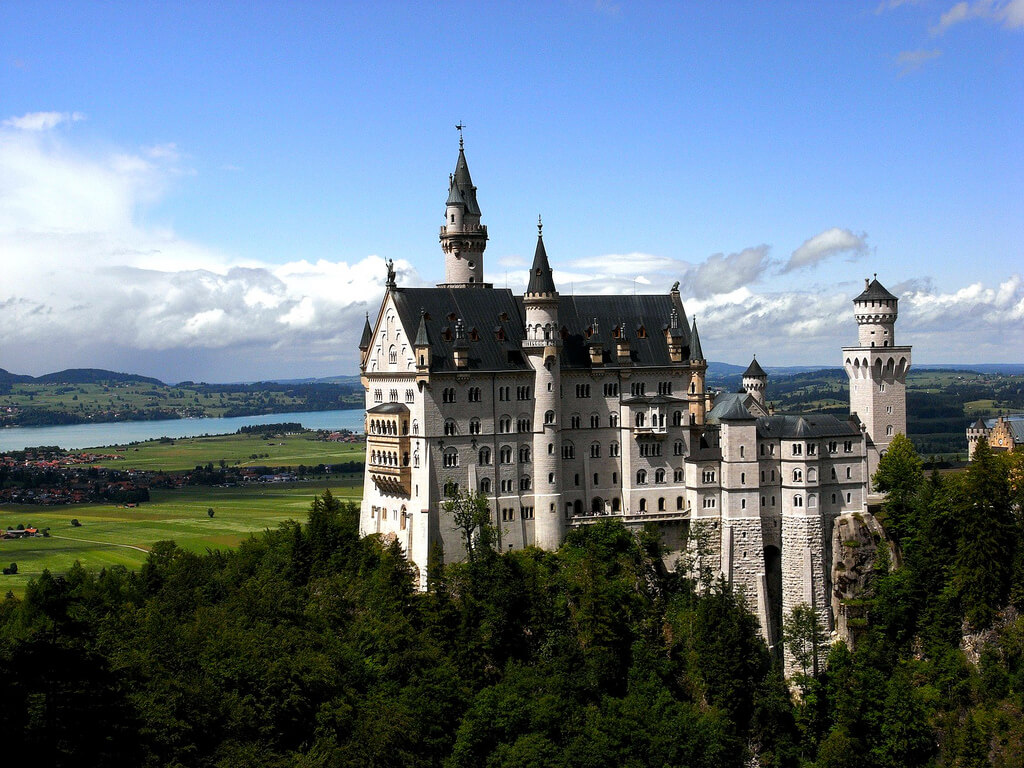}
    \end{minipage}
        & 
    \begin{minipage}{0.14\textwidth}
      \includegraphics[width=\textwidth]{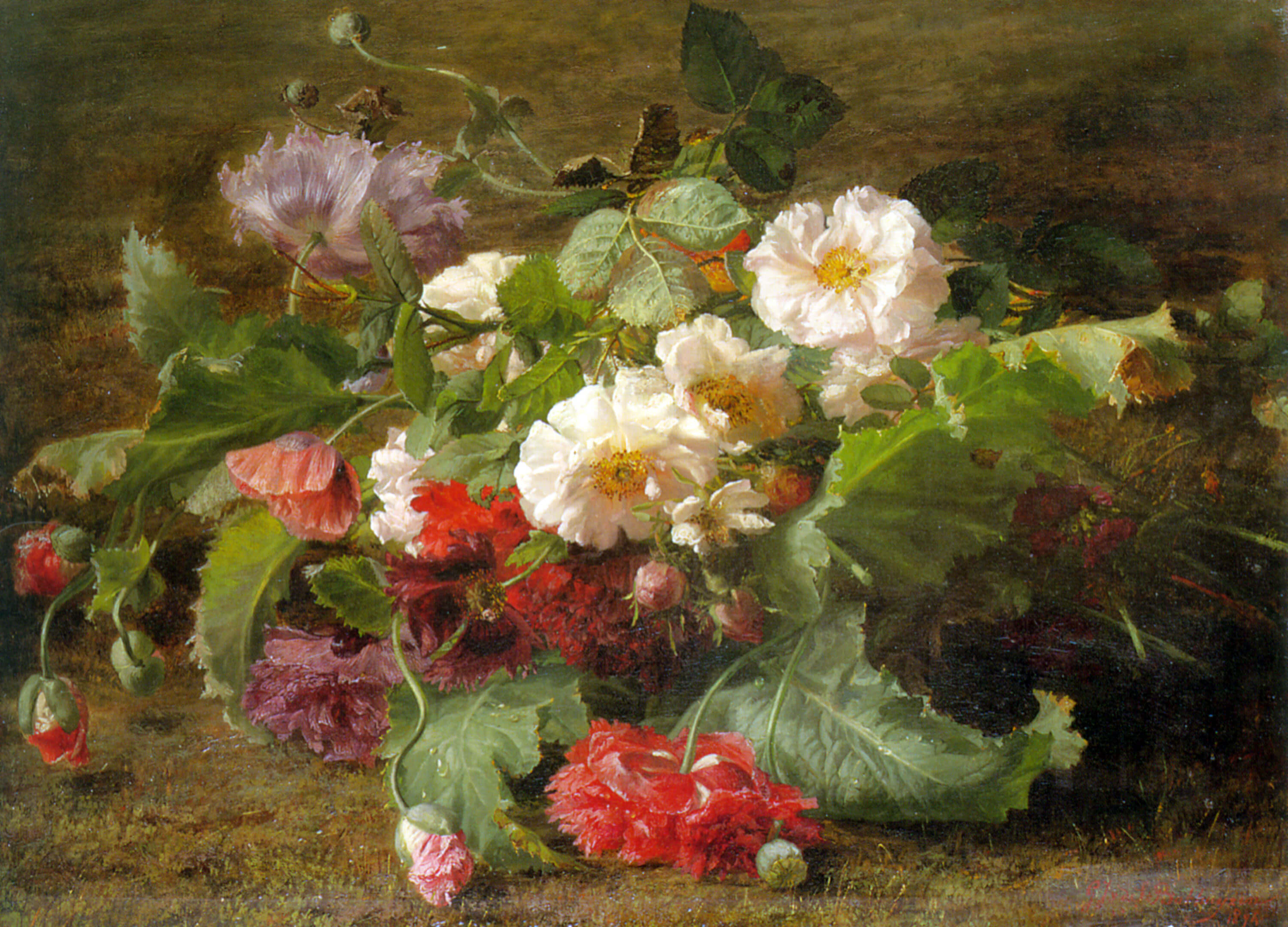}
    \end{minipage}
        & 
    \begin{minipage}{0.14\textwidth}
      \includegraphics[width=\textwidth]{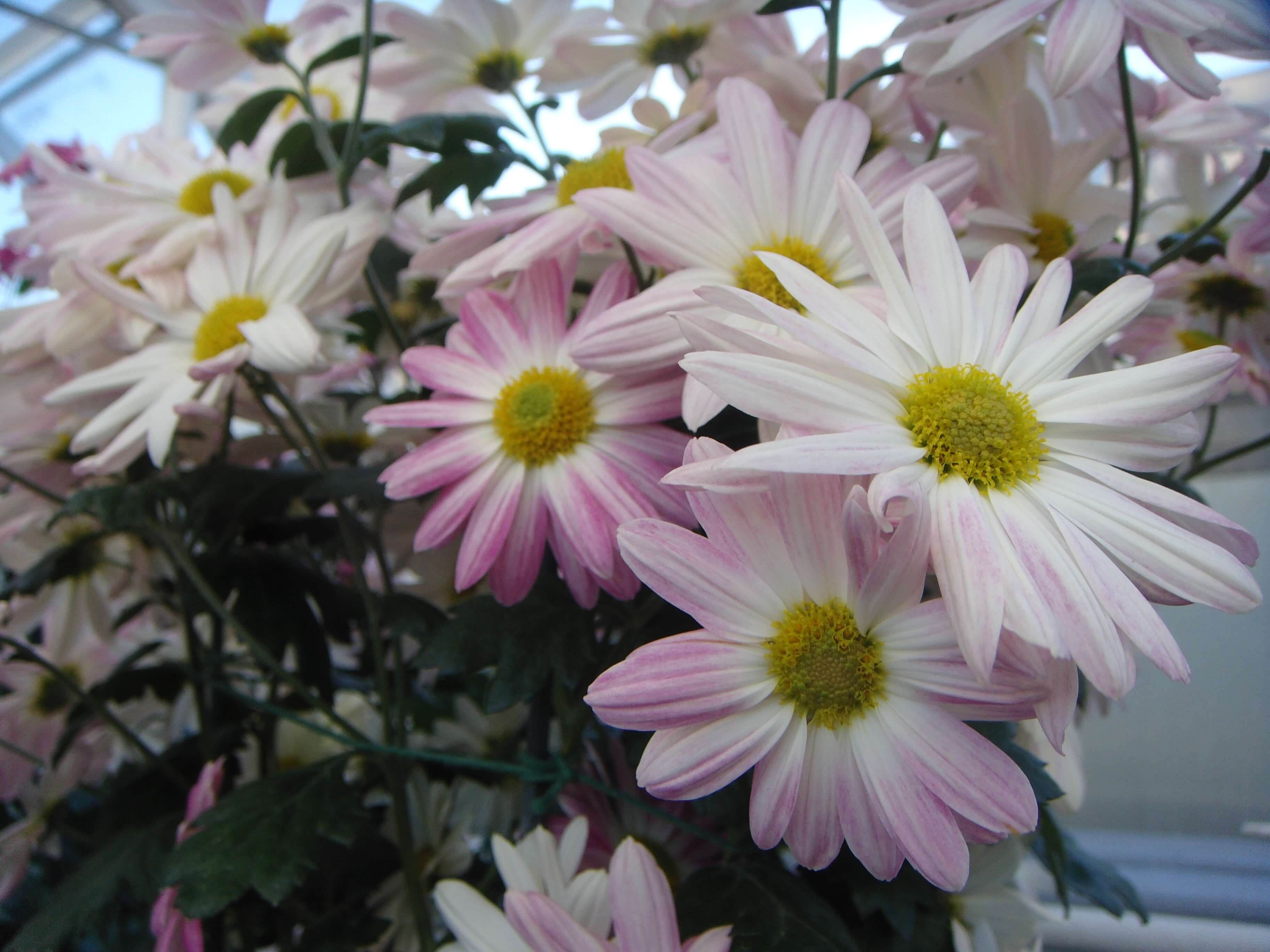}
    \end{minipage}
    \\
    \begin{minipage}{0.14\textwidth}
      \includegraphics[width=\textwidth]{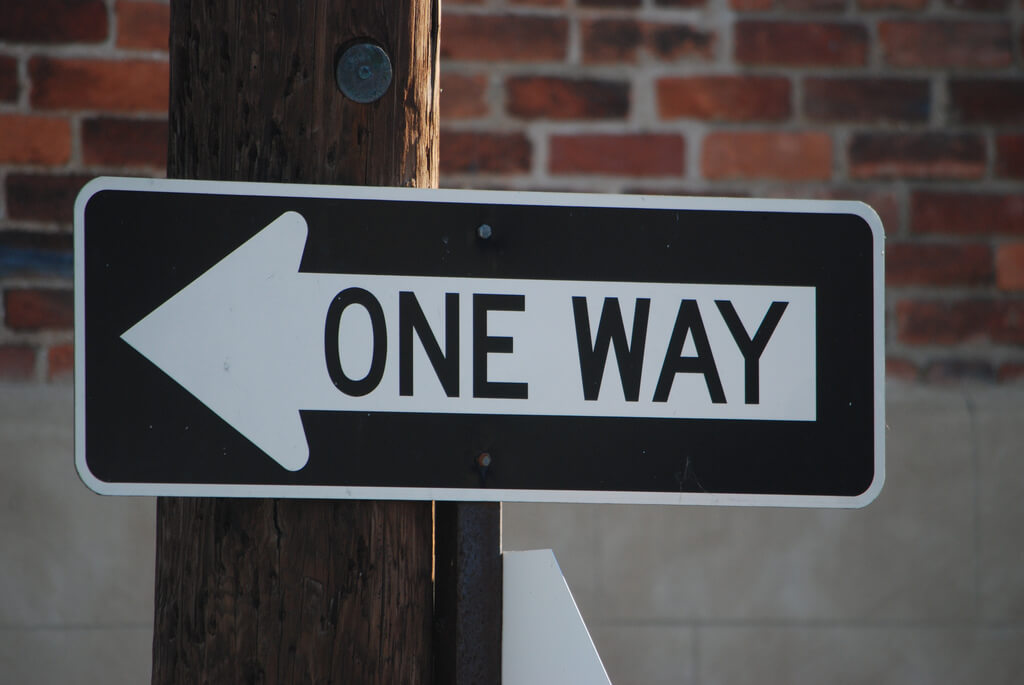}
    \end{minipage}
    &
    \begin{minipage}{0.14\textwidth}
      \includegraphics[width=\textwidth]{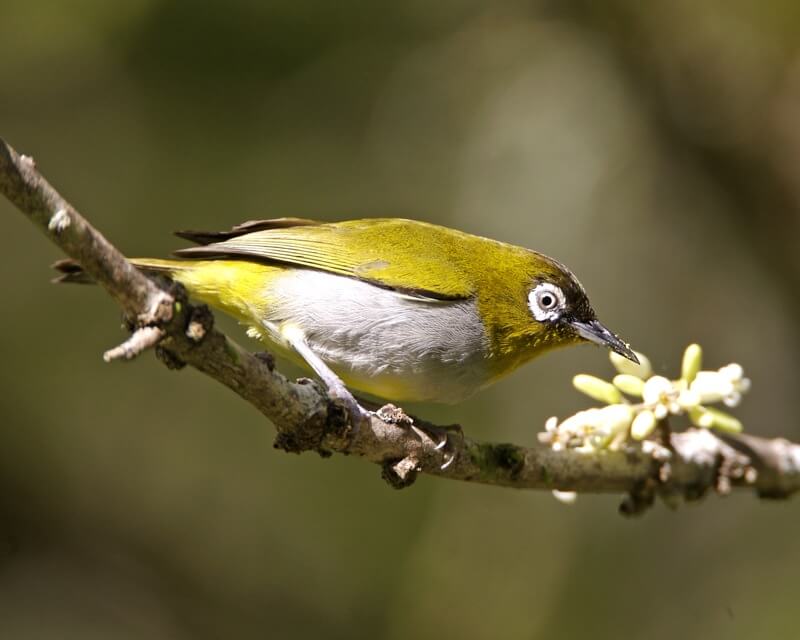}
    \end{minipage}
    & 
    \begin{minipage}{0.14\textwidth}
      \includegraphics[width=\textwidth]{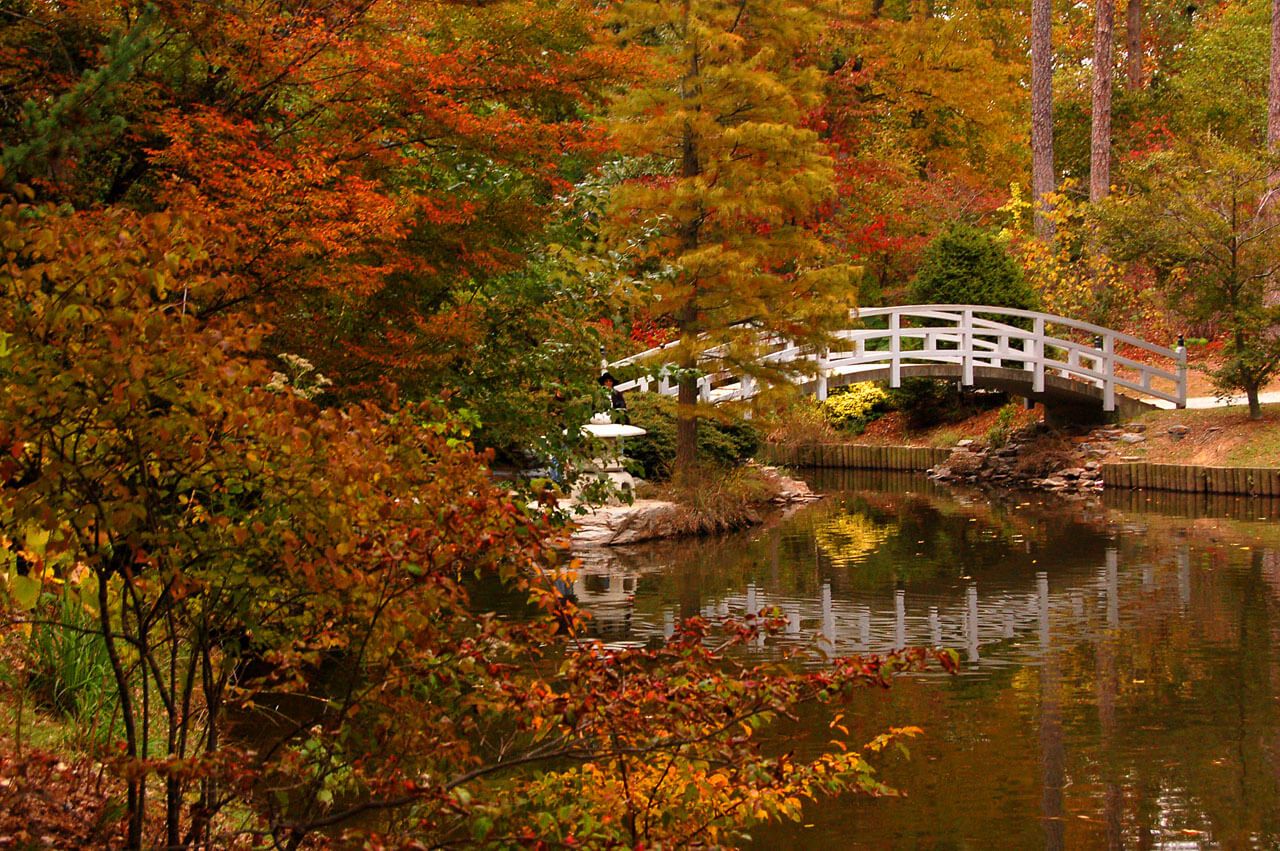}
    \end{minipage}
        & 
    \begin{minipage}{0.14\textwidth}
      \includegraphics[width=\textwidth]{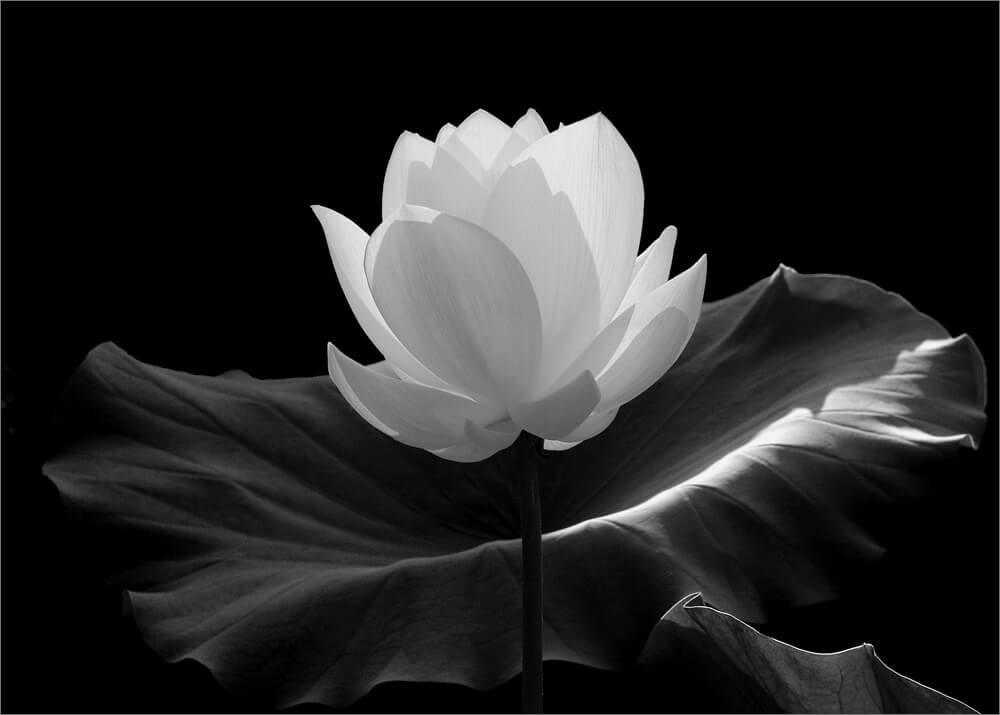}
    \end{minipage}
        & 
    \begin{minipage}{0.14\textwidth}
      \includegraphics[width=\textwidth]{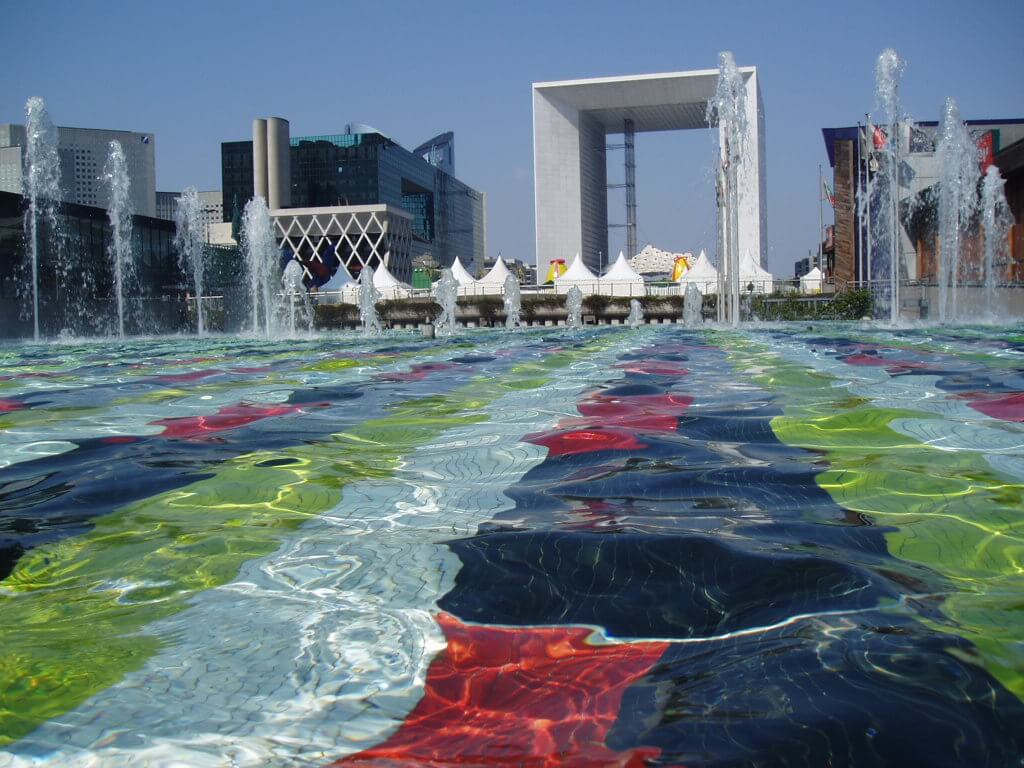}
    \end{minipage}
        & 
    \begin{minipage}{0.14\textwidth}
      \includegraphics[width=\textwidth]{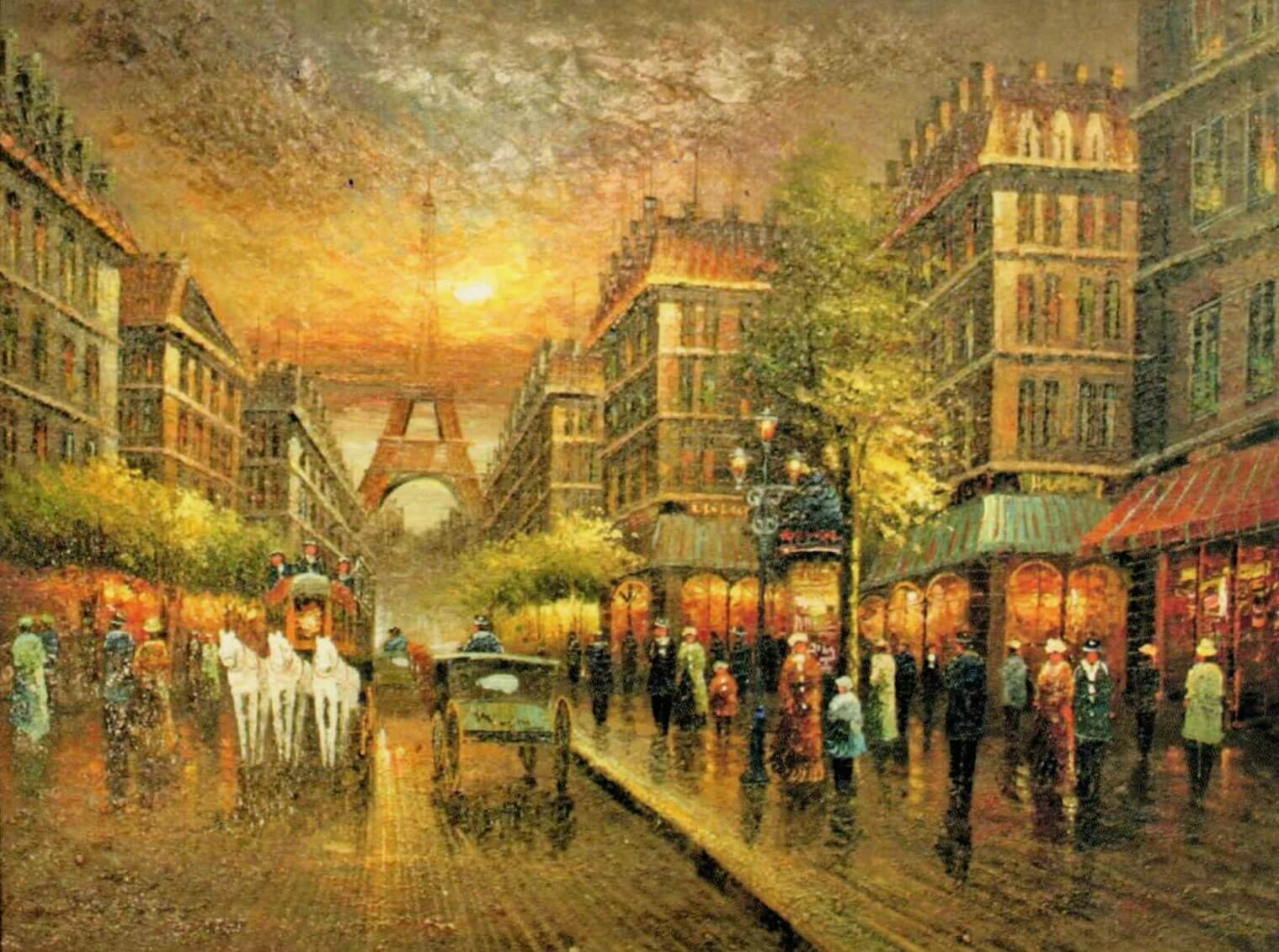}
    \end{minipage}
        & 
    \begin{minipage}{0.14\textwidth}
      \includegraphics[width=\textwidth]{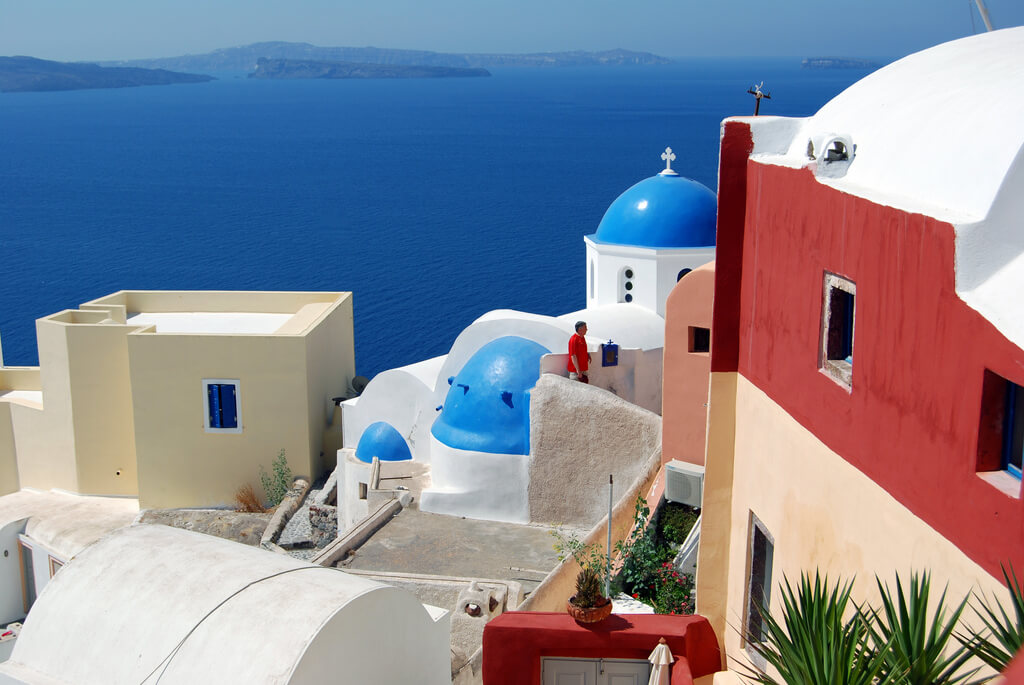}
    \end{minipage}
  \end{tabular}
  \vspace{5pt}
  \caption{Set of 14 images used to evaluate BER across several embedding algorithms and message step-sizes.}
 \label{tbl:test_images}
\end{table}

We compare the recovery error for several embedding algorithms across several step-sizes. Recovery error is defined as bit error rate (BER).
\[
\text{BER} = \frac{\text{count(incorrectly classified bits)}}{\text{count(all bits)}},
\]
A diverse set of 14 different images were used to test BER as shown in Figure \ref{tbl:test_images}. The camera used is a Basler acA2040-90uc-CVM4000, and the display used was a Acer S240HL LCD monitor.

\section{Results}

\begin{table}[!h]
\centering
  \begin{tabular}{ |l|r|r|r|r| }
   \hline \multicolumn{5}{ c }{Embedding Algorithms} \\ \hline
   \multicolumn{1}{|l|}{$|\delta|_2$}
   & \multicolumn{1}{l|}{Intensity}
   & \multicolumn{1}{l|}{Random}
   & \multicolumn{1}{l|}{\begin{tabular}[x]{@{}l@{}}$RGB$ \\ Differential \\ Metamers\end{tabular}}
   & \multicolumn{1}{l|}{\begin{tabular}[x]{@{}l@{}}$Lab$ \\ Differential \\ Metamers\end{tabular}} \\
   \specialrule{0.5pt}{0pt}{0pt}
  1 & 50.69\% & 50.99\% & 50.45\% & 49.85\% \\
  \specialrule{0.5pt}{0pt}{0pt}
  2 & 47.92\% & 48.81\% & 42.06\% & 42.06\% \\
  \specialrule{0.5pt}{0pt}{0pt}
  3 & 43.85\% & 46.97\% & 36.11\% & 37.25\% \\
  \specialrule{0.5pt}{0pt}{0pt}
  4 & 37.00\% & \cellcolor{red!15}44.59\% & 29.02\% & 27.83\% \\
  \specialrule{0.5pt}{0pt}{0pt}
  5 & \cellcolor{red!15}34.52\% & \cellcolor{red!15}42.41\% & 22.42\% & 21.73\% \\
  \specialrule{0.5pt}{0pt}{0pt}
  6 & \cellcolor{red!15}23.41\% & \cellcolor{red!15}41.22\% & \cellcolor{red!15}19.84\% & 17.61\% \\
  \specialrule{0.5pt}{0pt}{0pt}
  7 & \cellcolor{red!15}18.70\% & \cellcolor{red!15}38.10\% & \cellcolor{red!15}15.53\% & \cellcolor{red!15}15.08\% \\
  \specialrule{0.5pt}{0pt}{0pt}
  8 & \cellcolor{red!15}13.49\% & \cellcolor{red!15}35.57\% & \cellcolor{red!15}13.84\% & \cellcolor{red!15}12.80\% \\
  \specialrule{0.5pt}{0pt}{0pt}
  9 & \cellcolor{red!15}09.97\% & \cellcolor{red!15}34.72\% & \cellcolor{red!15}12.50\% & \cellcolor{red!15}12.00\% \\
  \specialrule{0.5pt}{0pt}{0pt}
  10 & \cellcolor{red!15}09.13\% & \cellcolor{red!15}32.89\% & \cellcolor{red!15}11.01\% & \cellcolor{red!15}10.91\% \\
  \specialrule{0.5pt}{0pt}{0pt}
  \end{tabular}
\vspace{5pt}
\caption{ BER for various embedding schemes (\textit{lower is better}). The red-shaded cells indicate step-sizes where an embedded, blended-checkerboard-pattern is easily visible. Differential metamers generated with trained ellipsoids in CIE $Lab$ are especially effective because both the BER is reduced and the threshold for acceptable step-size is increased.
Notice that for a mid-range step-size of 5 or 6, the color coding with differential metamers is significantly better than intensity modulation. }
\label{tab:ber_results_chart}
\end{table}

\begin{figure}[h!]
{
\hspace{1cm}
\vspace{-5pt}
\includegraphics[width=0.8\textwidth]{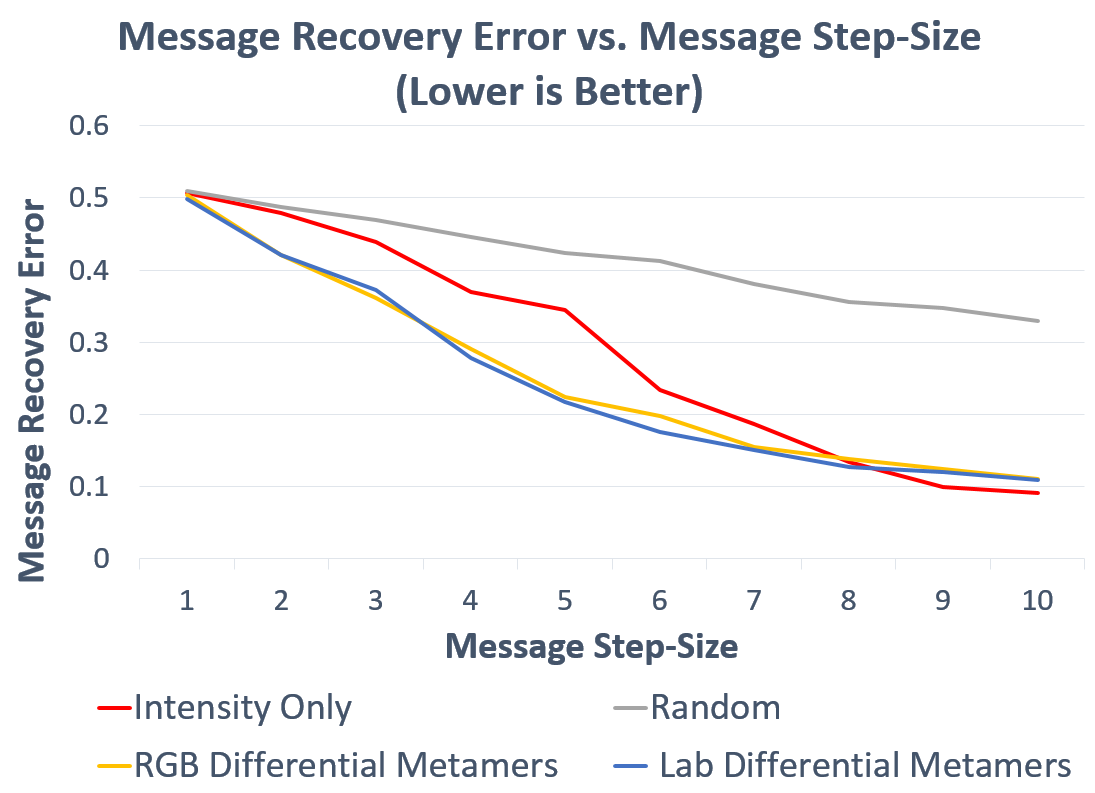}
}
\caption{ This graph compares message recovery across several embedding algorithms. Regardless of embedding algorithm, as the step-size increases, message recovery error decreases. However, large step-size also means a more visually obtrusive embedding. For an embedded message to be invisible, smaller step-sizes are greatly preferred. For small to mid-range step-sizes, color embedding using differential metamers is significantly better.
\label{fig:ber_results_graph}
}
\end{figure}

\begin{table*}[h]
\hspace{-0.55cm}
\subfloat[Low Texture Image]{
{\setlength{\extrarowheight}{4pt}%
  \begin{tabular}{ | c c | }
  \hline
  \multicolumn{2}{|c|}{Photographic Steganography:} \\ \multicolumn{2}{|c|}{Intensity vs Lab Differential Metamers} \\ \hline
     \multicolumn{1}{|c|}{Intensity}
   & \multicolumn{1}{c|}{Differential Metamers} \\
   \hline
   \multicolumn{2}{|c|}{Image with Embedded Message} \\
    \begin{minipage}{0.25\textwidth}
      \includegraphics[width=\textwidth]{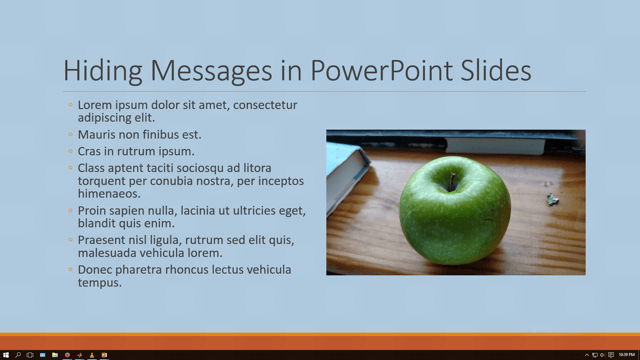}
    \end{minipage}
    &
    \begin{minipage}{0.25\textwidth}
      \includegraphics[width=\textwidth]{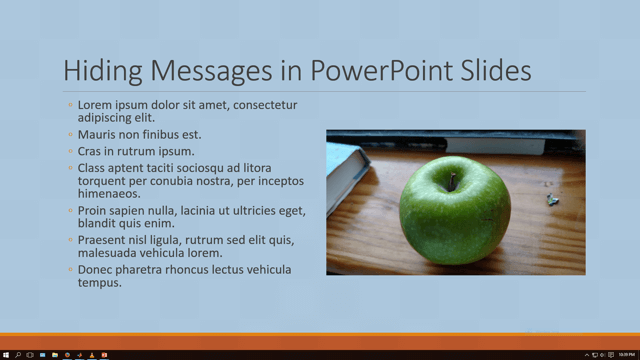}
    \end{minipage}
    \\
   \hline
   \multicolumn{2}{|c|}{Per-pixel difference} \\
    \begin{minipage}{0.25\textwidth}
      \includegraphics[width=\textwidth]{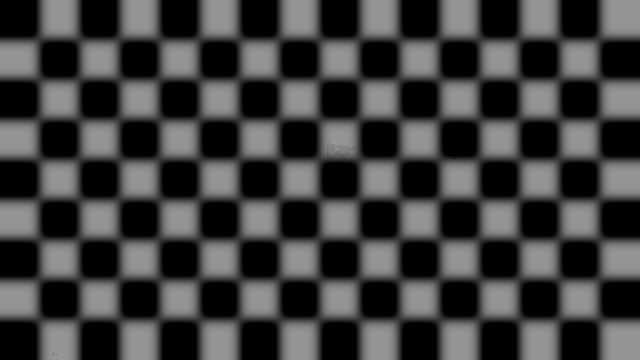}
    \end{minipage}
   	&
    \begin{minipage}{0.25\textwidth}
      \includegraphics[width=\textwidth]{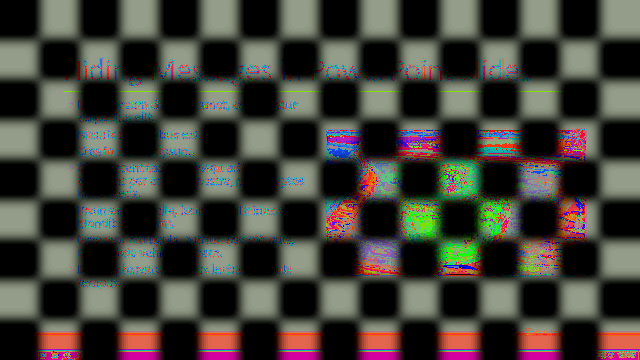}
    \end{minipage}
    \\
    \hline
   \multicolumn{2}{|c|}{Camera-recovered difference} \\
    \begin{minipage}{0.25\textwidth}
      \includegraphics[width=\textwidth]{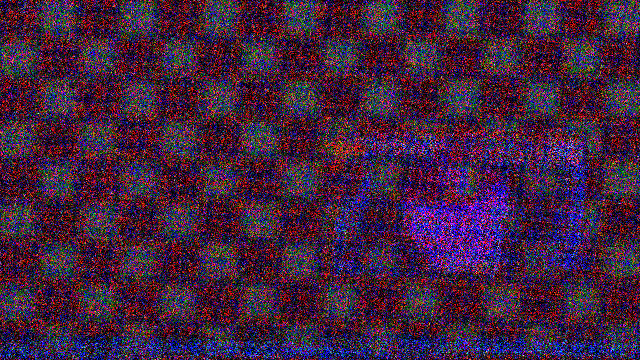}
    \end{minipage}
    & 
    \begin{minipage}{0.25\textwidth}
      \includegraphics[width=\textwidth]{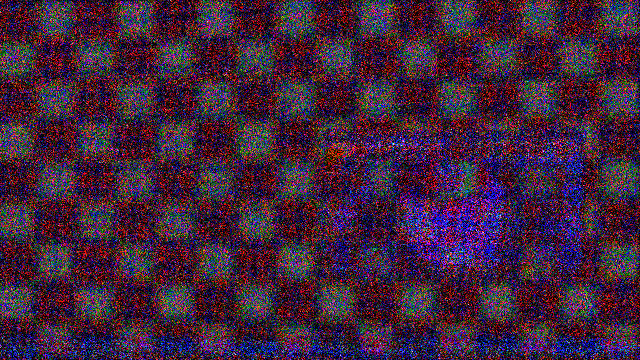}
    \end{minipage}
    \\
    \hline
   \multicolumn{2}{|c|}{Recovered Message} \\
    \begin{minipage}{0.25\textwidth}
      \includegraphics[width=\textwidth]{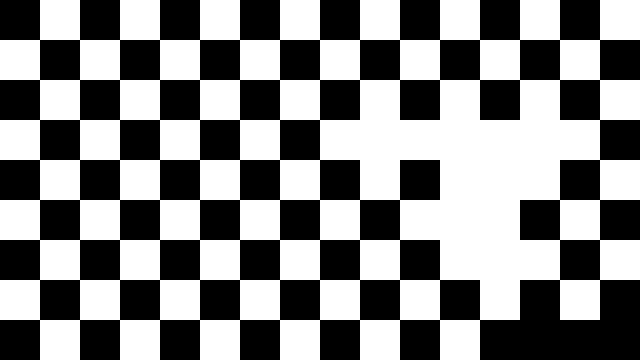}
    \end{minipage}
    & 
    \begin{minipage}{0.25\textwidth}
      \includegraphics[width=\textwidth]{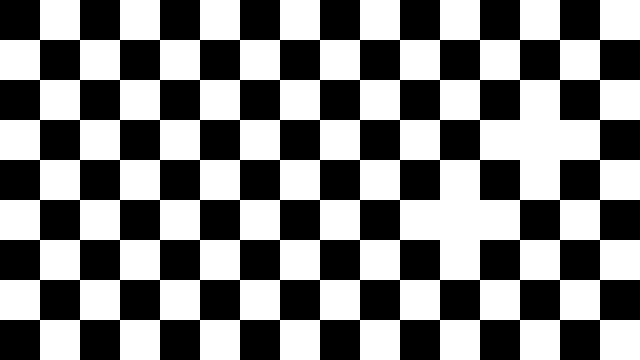}
    \end{minipage} 
    \\
    \hline
    \multicolumn{2}{|c|}{BER \textit{(lower is better)}} \\
    \hline
    \multicolumn{1}{|c|}{5.56\%}
    & \multicolumn{1}{c|}{1.39\%} \\
    \hline
  \end{tabular}}} 
\subfloat[Highly Textured Image]{
  {\setlength{\extrarowheight}{4pt}
  \begin{tabular}{ | c c | }
  \hline
  \multicolumn{2}{|c|}{Photographic Steganography:} \\ \multicolumn{2}{|c|}{Intensity vs Lab Differential Metamers} \\ \hline
     \multicolumn{1}{|c|}{Intensity}
   & \multicolumn{1}{c|}{Differential Metamers} \\
   \hline
   \multicolumn{2}{|c|}{Image with Embedded Message} \\
    \begin{minipage}{0.25\textwidth}
      \includegraphics[width=\textwidth]{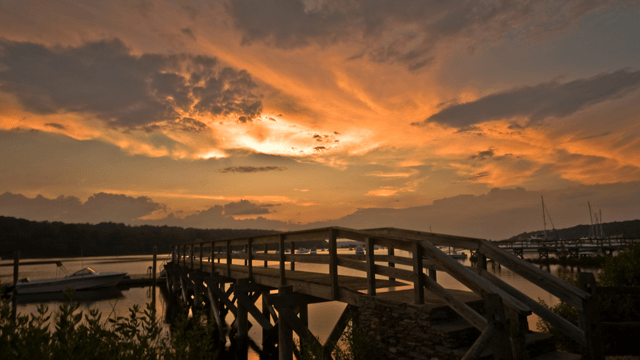}
    \end{minipage}
    &
    \begin{minipage}{0.25\textwidth}
      \includegraphics[width=\textwidth]{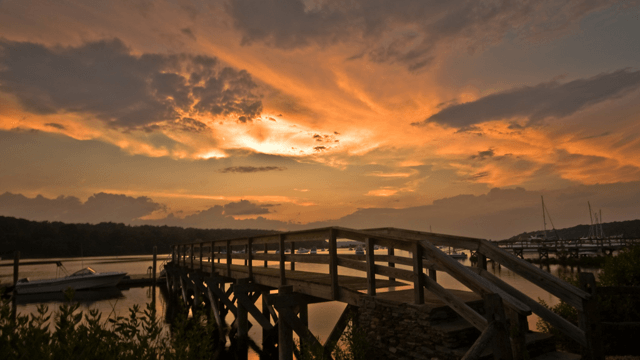}
    \end{minipage}
    \\
   \hline
   \multicolumn{2}{|c|}{Per-pixel difference} \\
    \begin{minipage}{0.25\textwidth}
      \includegraphics[width=\textwidth]{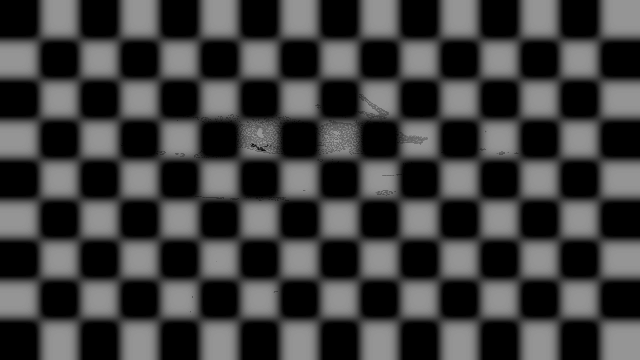}
    \end{minipage}
   	&
    \begin{minipage}{0.25\textwidth}
      \includegraphics[width=\textwidth]{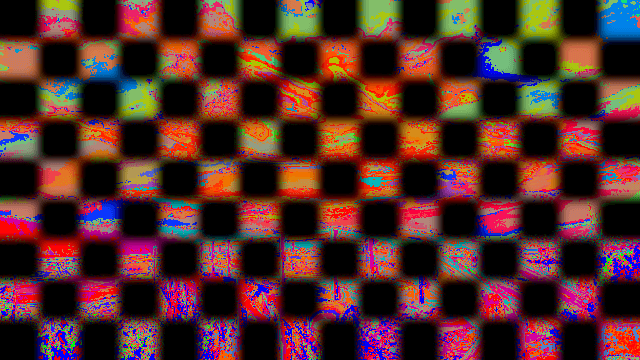}
    \end{minipage}
    \\
    \hline
   \multicolumn{2}{|c|}{Camera-recovered difference} \\
    \begin{minipage}{0.25\textwidth}
      \includegraphics[width=\textwidth]{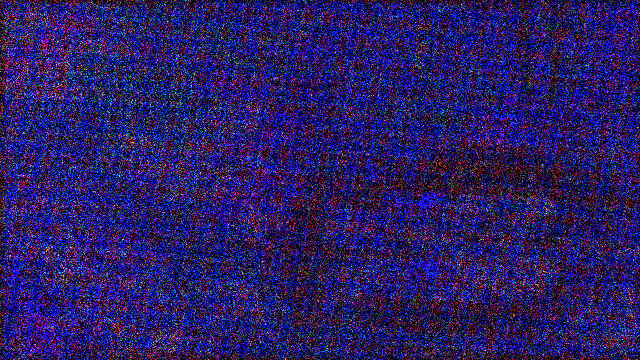}
    \end{minipage}
    & 
    \begin{minipage}{0.25\textwidth}
      \includegraphics[width=\textwidth]{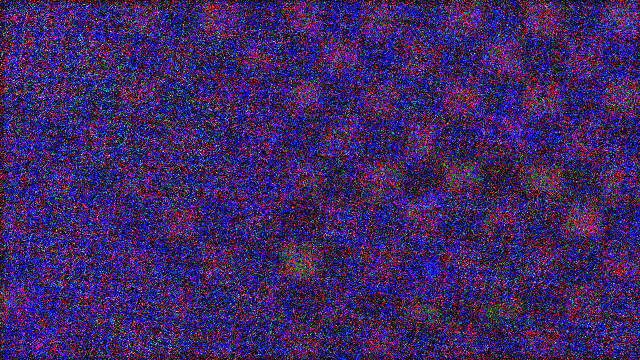}
    \end{minipage}
    \\
    \hline
   \multicolumn{2}{|c|}{Recovered Message} \\
    \begin{minipage}{0.25\textwidth}
      \includegraphics[width=\textwidth]{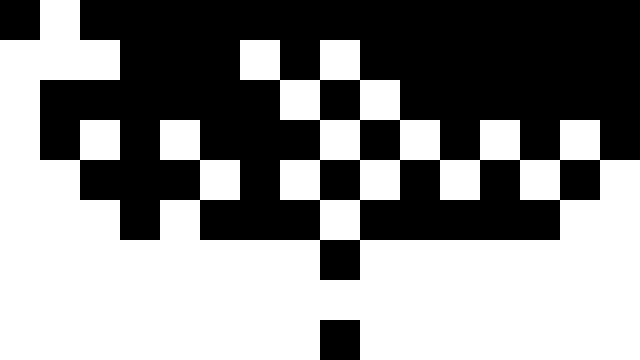}
    \end{minipage}
      & 
    \begin{minipage}{0.25\textwidth}
      \includegraphics[width=\textwidth]{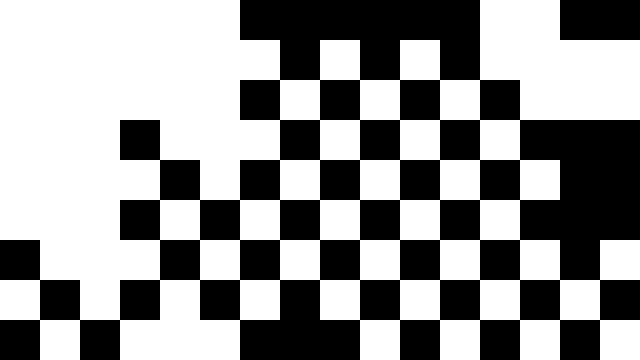}
    \end{minipage} 
    \\
    \hline
    \multicolumn{2}{|c|}{BER  \textit{(lower is better)}} \\
    \hline
      \multicolumn{1}{|c|}{38.19\%}
    & \multicolumn{1}{c|}{19.44\%} \\
    \hline
  \end{tabular}}}
  \vspace{5pt}
  \caption{Message embedding with intensity vs differential metamers example. The image in the first row is has been modulated to include a steganographic message. Below that, the Per-pixel difference shows the ground truth of exactly the changes that were made to the original image. The Camera-recovered difference shows the difference measured after the image has been displayed electronically, and captured by a camera. Notice that the differences between ground truth and camera-captured are large. Embedding messages with Lab differential metamers is effective for many types of images, including relatively flat images like powerpoint slides, as is shown in (a). The example in (b) showcases a more challenging case, where intensity embedding fails in dark and highly textured areas of the image. Lab differential metamers are significantly more effective for robust message embedding and recovery. In both (a) and (b), the $\delta$ is 5 across all algorithms.}
  \label{tbl:visual_results}
\end{table*}

To evaluate our embedding algorithms, a known message was embedded into a set of images. A camera then captured first the original image, then the image with the embedded message. Each algorithm was evaluated based on the accuracy of recovering each bit of the message. A wide range of message step-sizes were tested. Message step-size refers to the magnitude of the embedded message pixel values, which we refer to as $\delta$. A diverse set of 14 host images was used, shown in Table \ref{tbl:test_images}.
\\
\indent For the intensity-based approach, a uniform grayscale $\delta$ is applied to every pixel representing a ``1'' bit. The random approach applies a $\delta$ in a random direction to each pixel. The \text{RGB} differential metamers approach assigns a specialized $\delta$ value to each pixel in the base image. The differential metamer ellipsoids are trained in 6-dimensional \text{RGB} space. Similarly, the \textit{Lab} differential metamers approach assigns $\delta$ values from ellipsoids trained in \textit{Lab} space.
\\
\indent Table \ref{tab:ber_results_chart} shows the average message recovery for each embedding algorithm across a variety of $\delta$ step-sizes. The red-shaded cells represent values for which the step-size is $\delta$ is so large it can be easily seen. Figure \ref{fig:ber_results_graph} illustrates these results graphically.
\\
\indent For small step-sizes, the \text{RGB} and \textit{Lab} differential metamer approaches greatly outperform the alternatives. Small step-sizes are preferable because they are more difficult for humans to see. Differential metamers are designed to be less sensitive to human vision, but more sensitive to camera vision. So the approaches that incorporate differential metamers actually allow larger message step-sizes to be used. Differential metamers trained in perceptually uniform $Lab$ are highly effective at reducing human detection with more robust message recovery.
\\
\indent
Table \ref{tbl:visual_results} shows examples of photographic steganography where BER is significantly improved by embedding with differential metamers rather than intensity.


\section{Discussion and Conclusion}

\indent In this paper, we  present a color modulation method that can be used to steganographically embed data into images and videos.
We develop a data-driven approach to learn a pixel mapping function that produces a differential metamer pair for each input value.
These differential metamers are pairs of color values that minimize human visual response, but maximize camera response.
The key innovations is the exploitation of the mismatch between human spectral and camera sensitivity curves.
\\
\indent
We demonstrate the effectiveness of our differential metamer generation algorithm via message embedding.
The goal is to maximize throughput, minimize error, and reduce the visible effect to humans. Hidden camera-display messaging boasts a myriad of applications including: highly-directional networks, interactive museum exhibits, and indoor localization.

\section*{Acknowledgements}
This work was supported primarily by the National Science Foundation.
The Titan X used for this research was donated by the NVIDIA Corporation.

\clearpage
\bibliographystyle{splncs}
\bibliography{ColorEmbedding}

\end{document}